%% file: main.tex
\documentclass[11pt,a4paper]{article}
\usepackage[hyperref]{emnlp2020}

\aclfinalcopy 

\usepackage{microtype}
\usepackage{graphicx}
\usepackage{url}
\usepackage{subfigure}
\usepackage{amsmath}
\usepackage{booktabs} 

\usepackage{times}
\usepackage{latexsym}

\usepackage{balance}
\usepackage{url}
\usepackage{amssymb}
\usepackage{multirow}
\usepackage{makecell}
\usepackage{flushend}
\usepackage{color}
\usepackage{xcolor}
\usepackage{seqsplit}
\usepackage{caption}
\usepackage{courier}
\usepackage[russian,english]{babel}
\usepackage[T2A,T1]{fontenc}
\usepackage[utf8]{inputenc}

\usepackage[ruled,vlined]{algorithm2e}
\usepackage{algorithmic}

\title{Capturing document context inside sentence-level \\neural machine translation models with self-training}

\author{Elman Mansimov \\
  New York University \\
  {\tt mansimov@cs.nyu.edu} \\\And
  G\'abor Melis \\
  DeepMind \\
  {\tt melisgl@google.com} \\\And
  Lei Yu \\
  DeepMind \\
  {\tt leiyu@google.com}
  }

\date{}

\begin{document}
\maketitle
\begin{abstract}
Neural machine translation (NMT) has arguably achieved human level parity when trained and evaluated at the sentence-level. Document-level neural machine translation has received less attention and lags behind its sentence-level counterpart. The majority of the proposed document-level approaches investigate ways of conditioning the model on several source or target sentences to capture document context. These approaches require training a specialized NMT model from scratch on parallel document-level corpora. We propose an approach that doesn't require training a specialized model on parallel document-level corpora and is applied to a trained sentence-level NMT model at decoding time. We process the document from left to right multiple times and self-train the sentence-level model on pairs of source sentences and generated translations. Our approach reinforces the choices made by the model, thus making it more likely that the same choices will be made in other sentences in the document. We evaluate our approach on three document-level datasets: NIST Chinese-English, WMT'19 Chinese-English and OpenSubtitles English-Russian. We demonstrate that our approach has higher BLEU score and higher human preference than the baseline. Qualitative analysis of our approach shows that choices made by model are consistent across the document.
\end{abstract}

\section{Introduction}
Neural machine translation (NMT)~\citep{sutskever2014sequence,kalchbrenner2013recurrent,bahdanau2014neural} has achieved great success, arguably reaching the levels of human parity~\citep{hassan2018zhenparity} on Chinese to English news translation that led to its popularity and adoption in academia and industry. These models are predominantly trained and evaluated on sentence-level parallel corpora. Document-level machine translation that requires capturing the context to accurately translate sentences has been recently gaining more popularity and was selected as one of the main tasks in the premier machine translation conference WMT19~\citep{barrault2019wmt}. 

A straightforward solution to translate documents by translating sentences in isolation leads to inconsistent but syntactically valid text. The inconsistency is the result of the model not being able to resolve ambiguity with consistent choices across the document. For example, the recent NMT system that achieved human parity~\cite{hassan2018zhenparity} inconsistently used three different names "Twitter Move Car", "WeChat mobile", "WeChat move" when referring to the same entity~\citep{sennrichdoc}.

To tackle this issue, the majority of the previous approaches~\citep{jean2017context,wang2017exploitcorss,kuang2017coherencenmt,tiedemann2017neuralcontext,maruf2018document,agrawal2018context,zhang2018doctransformer,xiong2018deliberationdoc,miculicich2018document,voita2019monocontext,voita2019context,jean2019fillin,marcin2019docnmt} proposed context-conditional NMT models trained on document-level data. However, none of the previous approaches are able to exploit trained NMT models on sentence-level parallel corpora and require training specialized context-conditional NMT models for document-level machine translation.

We propose a way of incorporating context into a trained sentence-level neural machine translation model at decoding time. We process each document monotonically from left to right one sentence at a time and self-train the sentence-level NMT model on its own generated translation. This procedure reinforces choices made by the model and hence increases the chance of making the same choices in the remaining sentences in the document. Our approach does not require training a separate context-conditional model on parallel document-level data and allows us to capture context in documents using a trained sentence-level model.  

We make the key contribution in the paper by introducing the first document-level neural machine translation approach that does not require training a context-conditional model on document data. We show how to adapt a trained sentence-level neural machine translation model to capture context in the document during decoding.
We evaluate and demonstrate improvements of our proposed approach measured by BLEU score and preferences of human annotators on several document-level machine translation tasks including NIST Chinese-English, WMT19 Chinese-English and OpenSubtitles English-Russian datasets. We qualitatively analyze the decoded sentences produced using our approach and show that they indeed capture the context.

\section{Proposed Approach}

\begin{algorithm}[t]
   \caption{Document-level NMT with self-training at decoding time}
   \label{alg:selftrain}
\SetAlgoLined
\begin{algorithmic}
\STATE {\bfseries Input:} Document $\boldsymbol{D} = (X_1, ..., X_n)$, pretrained sentence-level NMT model $f(\theta)$, learning rate $\alpha$ and decay prior $\lambda$
\STATE {\bfseries Output:} Translated sentences $(Y_1, ..., Y_n)$
\STATE Backup original values of parameters $\tilde{\theta} \leftarrow \theta$
\FOR{$i=1$ {\bfseries to} $n$}
\STATE Translate sentence $X_i$ using sentence-level model $f(\theta)$ into target sentence $Y_i$.
\STATE Calculate cross-entropy loss $\mathit{L}(X_{i},Y_{i})$ \\ using $Y_i$ as target.
\FOR{$j=1$ {\bfseries to} $m$}
\STATE $\theta \leftarrow \theta - \alpha \nabla_{\theta} \mathit{L}(X_{i}, Y_{i}) + \lambda (\tilde{\theta} - \theta) $
\ENDFOR
\ENDFOR
\end{algorithmic}
\end{algorithm}

We translate a document $\boldsymbol{D}$ consisting of $n$ source sentences $X_{1}, X_{2}, ..., X_{n}$ into the target language, given a well-trained sentence-level neural machine translation model $f_{\theta}$. The sentence-level model parametrizes a conditional distribution $p(Y|X) = \prod_{i=1}^{T} p(y_t|Y_{<t},X)$ of each target word $y_{t}$ given the preceding words $Y_{<t}$ and the source sentence $X$. Decoding is done by approximately finding $\arg \max_{Y} p(Y|X)$ using greedy decoding or beam-search. $f$ is typically a recurrent neural network with attention~\citep{bahdanau2014neural} or a Transformer model~\citep{vaswani2017attention} with parameters $\theta$.

\subsection{Self-training} \label{subsec:dyneval}
We start by translating a first source sentence $X_{1}$ in the document $\boldsymbol{D}$ into the target sentence $Y_{1}$. We then self-train the model on the sentence pair $(X_{1},Y_{1})$, which maximizes the log probabilities of each word in the generated sentence $Y_{1}$ given source sentence $X_{1}$. The self-training procedure runs gradient descent steps for a fixed number of steps with a weight decay. Weight decay keeps the updated values of weights closer to original values. We repeat the same update process for the remaining sentences in the document. The detailed implementation of self-training procedure during decoding is shown in Algorithm~\ref{alg:selftrain}.

\subsection{Multi-pass self-training}
Since the document is processed in the left-to-right, monotonic order, our self-training procedure does not incorporate the choices of the model yet to be made on unprocessed sentences. In order to leverage global information from the full document and to further reinforce the choices made by the model across all generated sentences, we propose multi-pass document decoding with self-training. Specifically, we process the document multiple times monotonically from left to right while continuing self-training of the model.

\subsection{Oracle self-training to upper bound performance}

Since generated sentences are likely to contain some errors, our self-training procedure can reinforce those errors and thus potentially hurt the performance of the model on unprocessed sentences in the document. In order to isolate the effect of imperfect translations and estimate the upper bound of performance, we evaluate our self-training procedure with ground-truth translations as targets, which we call \emph{oracle self-training}. Running oracle self-training makes it similar to the dynamic evaluation approach introduced in language modeling~\citep{mikolov2012phd,graves2013seq,krause2018dyneval}, where input text to the language model is the target used to train the neural language model during evaluation. We do not use the oracle in multi-pass self-training since this would make it equivalent to memorizing the correct translation for each sentence in the document and regenerating it again.

\section{Related Work}

Although there have been some attempts at tackling document-level neural machine translation (for example see proceedings of discourse in machine translation workshop~\citep{discomt}), it has largely received less attention compared to sentence-level neural machine translation. Prior document-level NMT approaches~\citep{jean2017context,wang2017exploitcorss,kuang2017coherencenmt,tiedemann2017neuralcontext,maruf2018document,agrawal2018context,zhang2018doctransformer,miculicich2018document} proposed different ways of conditioning NMT models on several source sentences in the document. Perhaps closest of those document NMT approaches to our work is the approach by ~\cite{kuang2017coherencenmt}, where they train a NMT model with a separate non-parametric cache~\citep{kuhn1990cache} that incorporates topic information about the document. Recent approaches~\citep{jean2019fillin,marcin2019docnmt,voita2019monocontext} use only partially available parallel document data or monolingual document data. These approaches proposed to fill in missing context in the documents with random or generated sentences. Another line of document-level NMT work~\citep{xiong2018deliberationdoc,voita2019context} proposed a two-pass document decoding model inspired by the deliberation network~\citep{xia2017deliberation} in order to incorporate target side document context. Recently, \citet{yu2019docnoisychannel} proposed a novel beam search method that incorporates document context inside noisy channel model~\citep{shannon1948noisychannel,yu2017noisychannel,yee2019noisychannel}. Similar to our work, their approach doesn't require training context-conditional models on parallel document corpora, but relies on separate target-to-source NMT model and unconditional language model to re-rank hypotheses of the source-to-target NMT model.

Closest to our work is the \textit{dynamic evaluation} approach proposed by \cite{mikolov2012phd} and further extended by \cite{graves2013seq,krause2018dyneval}, where a neural language model is trained at evaluation time. However unlike language modeling where inputs are ground-truth targets used both during training and evaluation, in machine translation ground-truth translation are not available at decoding time in practical settings. The general idea of storing memories in the weights of the neural network rather than storing memories as copies of neural network activations, that is behind our approach and dynamic evaluation, goes back to 1970s and 1980s work on associative memory models~\citep{willshaw1969associative,kohonen1972associative,anderson1981associative,hopfield1982associative} and to more recent work on fast weights~\citep{ba2016fastweights}.

Our work belongs to the broad category of self-training or pseudo-labelling approaches~\citep{scudder1965selftrain,lee2013pseudolabel} proposed to annotate the unlabeled data to train supervised classifiers. Self-training has been successfully applied to NLP tasks such as word-sense disambiguation~\citep{yarowsky1995word} and parsing~\citep{mcclosky2006selftrain,reichart2007selftrain,huang2009selftrain}. Self-training has also been used to label monolingual data to improve the performance of sentence-level statistical and neural machine translation models~\citep{ueffing2006selfmt,zhang2016selfmt}. Recently, \citet{he2019selftrain} proposed noisy version of self-training and showed improvement over classical self-training on machine translation and text summarization tasks. Backtranslation~\citep{sennrich2016bt} is another popular pseudo-labelling technique that utilizes target-side monolingual data to improve performance of NMT models.

\section{Experiments}

\begin{table*}[t!]
\centering
\input{tables/nist-results.tex}
\label{tab:nistresults}
\end{table*}

\begin{table*}[t!]
\centering
\input{tables/nist-ablation-results.tex}
\label{tab:nistablate}
\end{table*}

\subsection{Datasets}
We use the NIST Chinese-English (Zh-En), the WMT19 Chinese-English (Zh-En) and the OpenSubtitles English-Russian (En-Ru) datasets in our experiments. 

The NIST training set consists of 1.5M sentence pairs from LDC-distributed news. We use MT06 set as validation set. We use MT03, MT04, MT05 and MT08 sets as held out test sets. The MT06 validation set consists of $1649$ sentences with $21$ sentences per document. MT03, MT04, MT05 and MT08 consist of $919$, $1788$, $1082$ and $1357$ sentences with $9$, $9$, $11$ and $13$ sentences on average per document respectively. We follow previous work~\citep{zhang2018doctransformer} when preprocessing NIST dataset. We preprocess the NIST dataset with punctuation normalization, tokenization, and lower-casing. Sentences are encoded using byte-pair encoding~\citep{sennrich2016subwords} with source and target vocabularies of roughly 32K tokens. We use the case-insensitive {\tt multi-bleu.perl} script with $4$ reference files to evaluate the model.

The WMT19 dataset includes the UN corpus, CWMT, and news
commentary. We filter the training data by removing duplicate sentences and sentences longer than 250 words. The training dataset consits of 18M sentence pairs. We use newsdev2017 as a validation set and use newstest2017, newstest2018 and newstest2019 as held out test sets. newsdev2017, newstest2017, newstest2018 and newstest2019 consist of total of $2002$, $2001$, $3981$ and $2000$ sentences with average of $14$, $12$, $15$ and $12$ sentences per document respectively. We similarly follow previous work~\citep{xia2019wmtmicrosoft} when preprocessing the dataset. Chinese sentences are preprocessed by segmenting and normalizing punctuation. English sentences are preprocessed by tokenizing and true casing. We learn a byte-pair encoding~\citep{sennrich2016subwords} with source and target vocabularies of roughly 32K tokens. We use sacreBLEU~\citep{post2018sacrebleu} for evaluation.

The OpenSubtitles English-Russian dataset, consisting of movie and TV subtitles, was prepared by \cite{voita2019context}.\footnote{\url{https://github.com/lena-voita/good-translation-wrong-in-context}} The training dataset consists of 6M parallel sentence pairs. We use the context aware sets provided by the authors consisting of $10000$ documents both in validation and test sets. Due to the way the dataset is processed, each document only contains $4$ sentences. The dataset is preprocessed by tokenizing and lower casing. We use byte-pair encoding~\citep{sennrich2016subwords} to prepare source and target vocabularies of roughly 32K tokens. We use {\tt multi-bleu.perl} script for evaluation.

\subsection{Hyperparameters}

\begin{table*}[t!]
\centering
\input{tables/wmt-results.tex}
\label{tab:wmtresults}
\end{table*}

\begin{table}[t!]
\centering
\input{tables/opus-results.tex}
\label{tab:opusresults}
\end{table}

We train a Transformer~\citep{vaswani2017attention} on all datasets. Following previous~\citep{zhang2018doctransformer,voita2019context,xia2019wmtmicrosoft} work we use the Transformer base configuration ({\tt transformer\_base}) on the NIST Zh-En and the OpenSubtitles En-Ru datasets and use the Transformer big configuration ({\tt transformer\_big}) on the WMT19 Zh-En dataset. Transformer base consists of $6$ layers, $512$ hidden units and $8$ attention heads. Transformer big consists of $6$ layers, $1024$ hidden units and $16$ attention heads. We use a dropout rate~\citep{srivastava2014dropout} of $0.1$ and label smoothing to regularize our models. We train our models with the Adam optimizer~\citep{kingma2014adam} using the same warm-up learning rate schedule as in \cite{vaswani2017attention}. During decoding we use beam search with beam size $4$ and length penalty $0.6$. We additionally train backtranslated models~\citep{sennrich2016bt} on the NIST Zh-En and the OpenSubtitles En-Ru datasets. We use the publicly available English gigaword dataset~\citep{gigaword} to create synthetic parallel data for the NIST Zh-En dataset and use synthetic parallel data provided by \citep{voita2019monocontext} for the OpenSubtitles En-Ru dataset. When training backtranslated models, we oversample the original parallel data to make the ratio of synthetic data to original data equal to $1$~\citep{edunov2018backtranslate}. We tune the number of update steps, learning rate, decay rate, and number of passes over the document of our self-training approach with a random search on a validation set. We use the range of $(5 \times 10^{-5}, 5 \times 10^{-1})$ for learning rate, range of $(0.001, 0.999)$ for decay rate, number of update steps ($2, 4, 8$) and number of passes over the document ($2, 4$) for random search. We found that best performing models required a small number of update steps (either $2$ or $4$) with a relatively large learning rate ($\sim 0.005-0.01)$ and small decay rate ($\sim 0.2-0.5$). We use the Tensor2Tensor library~\citep{tensor2tensor} to train baseline models and to implement our method.

\section{Results}
\begin{table*}[t]
\centering
\begin{minipage}{0.95\columnwidth}
\centering
\includegraphics[width=0.99\columnwidth]{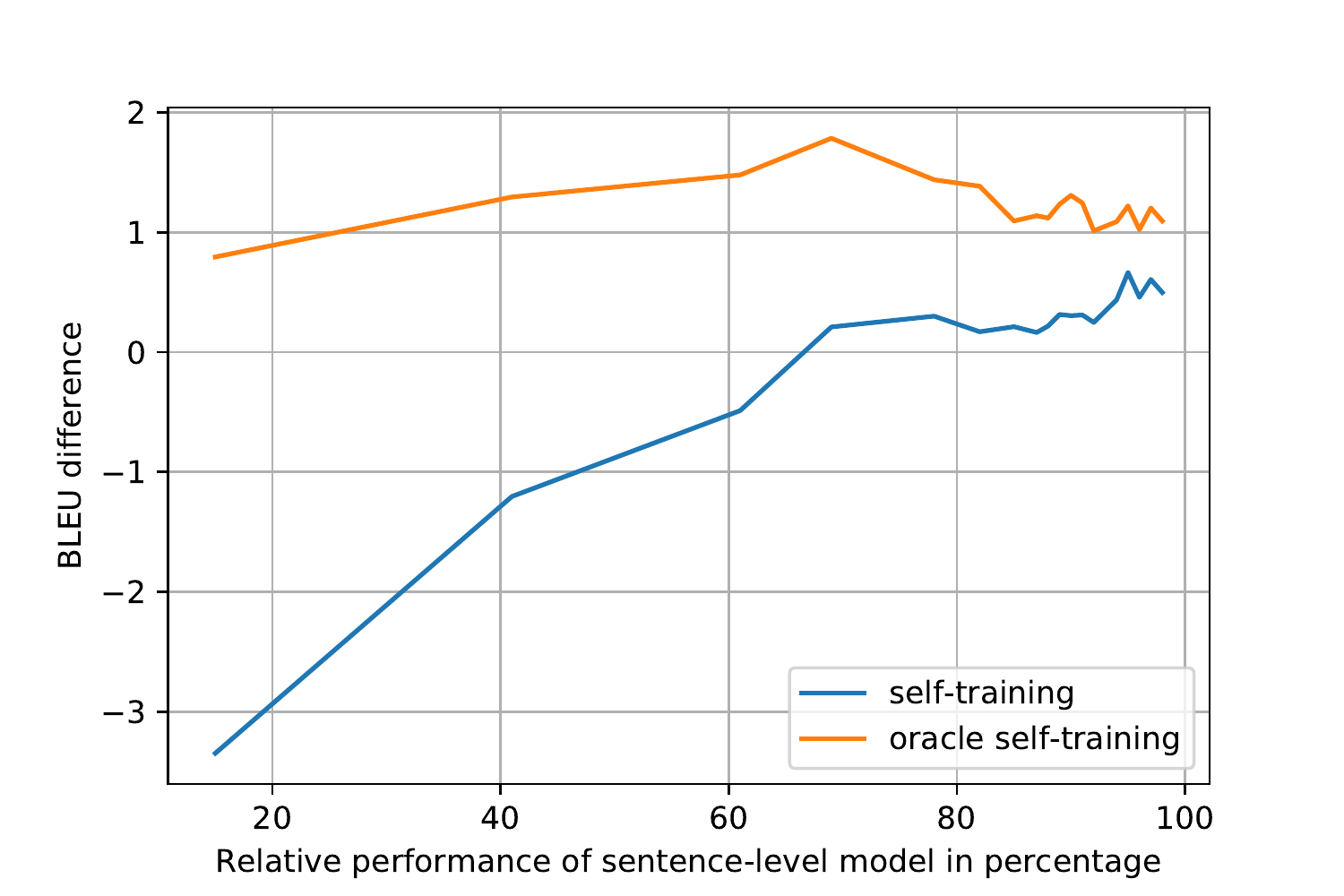}
\captionof{figure}{Relationship between relative performance of the sentence-level model and BLEU difference of self-training on the NIST dataset.}
\label{fig:ckptablate}
\end{minipage}
\quad\quad\quad
\begin{minipage}{0.95\columnwidth}
\centering
\includegraphics[width=0.99\columnwidth]{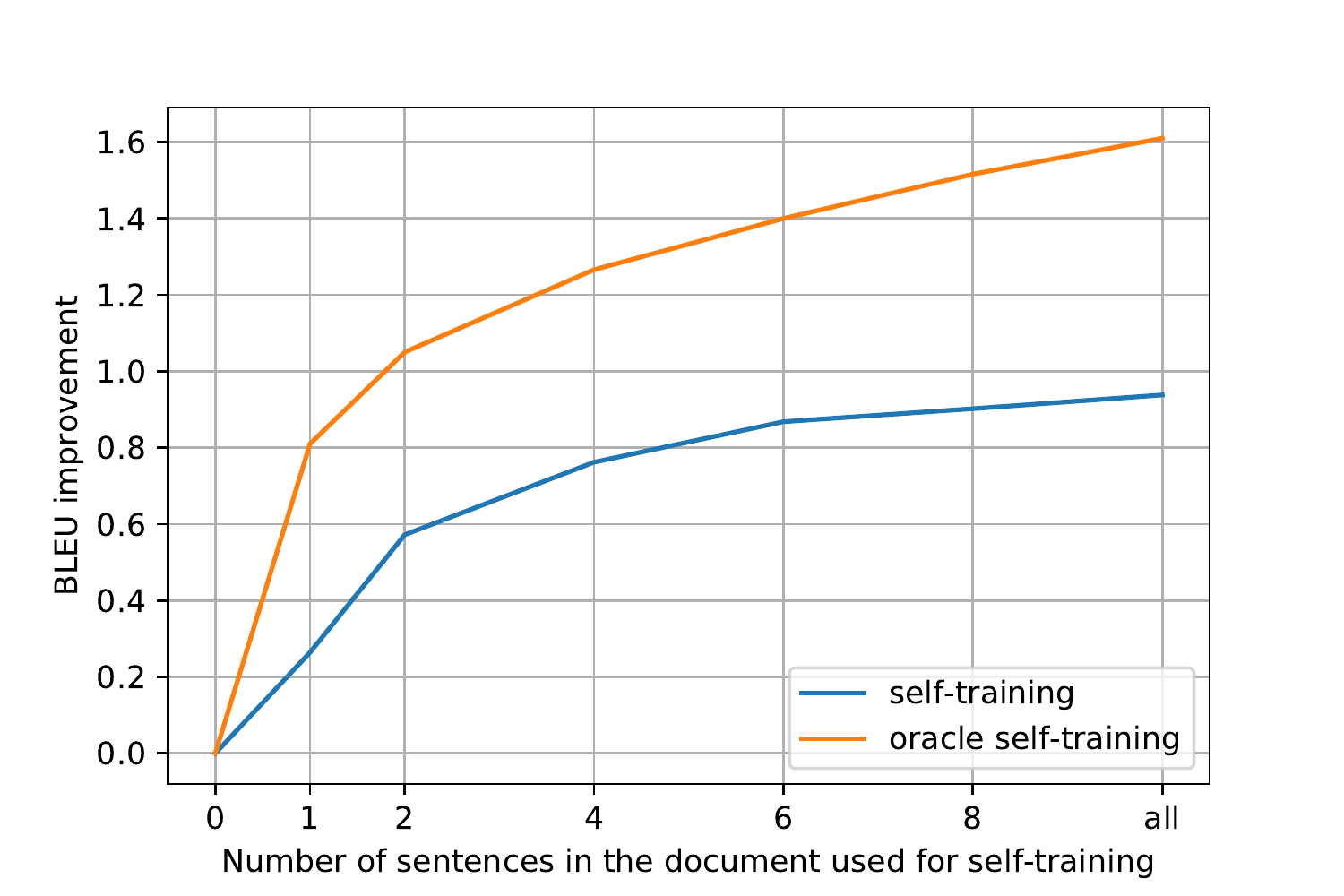}
\captionof{figure}{Relationship between number of sentences and BLEU improvement of self-training on the NIST dataset.}
\label{fig:numsent}
\end{minipage}
\end{table*}

\begin{table}[h]
\centering
\input{tables/nist-human-eval.tex}
\caption{Human evaluation results on the NIST Zh-En and the OpenSubtitles En-Ru datasets. "Total" denotes total number of annotations collected from humans. "Self-train" denotes number of times evaluators preferred documents by the self-training approach. "Baseline" denotes number of times evaluators preferred documents by sentence-level baseline.}
\label{tab:nist_human_eval}
\end{table}

We present translation quality results measured by BLEU on NIST dataset on Table~\ref{tab:nistresults}. The self-training procedure improves the results of our sentence-level baseline by the average of $0.53$ BLEU for non-backtranslated model and by $0.93$ BLEU for backtranslated model for all evaluation sets. Our baseline sentence-level Transformer model trained without backtranslation outperforms previous document-level models by \citet{wang2017exploitcorss} and \citet{kuang2017coherencenmt} and is comparable to the document-level model proposed by ~\citet{zhang2018doctransformer}. Backtranslation further improves the results of our sentence-level model leading to higher BLEU score compared to the Document Transformer~\cite{zhang2018doctransformer}.

In Table~\ref{tab:nistablate}, we show a detailed study of effects of multi-pass self-training and oracle self-training on BLEU scores on NIST evaluation sets. First, multiple decoding passes over the document give an additional average improvement of $0.25-0.45$ BLEU points compared to the single decoding pass over the document. Using oracle self-training procedure gives an average of $0.86$ and $1.63$ BLEU improvement over our non-backtranslated and backtranslated sentence-level baseline models respectively. Compared to using generated translations by the model, oracle self-training gives an improvement of $0.3$ and $0.7$ BLEU points for non-backtranslated and backtranslated models respectively.

The results on the WMT19 evaluation sets are presented on Table~\ref{tab:wmtresults}. Compared to the NIST dataset our self-training procedure shows an improvement of $0.1$ BLEU over a sentence-level baseline model. Oracle self-training outperforms sentence-level baselines by a significant margin of $2.5$ BLEU. We hypothesize that such a large gap between performance of oracle and non-oracle self-training is due to the more challenging nature of the WMT dataset which is reflected in the worse performance of sentence-level baseline on WMT compared to NIST. We investigate this claim by measuring the relationship between BLEU achieved by self-training and the relative quality of the sentence-level model on the NIST dataset. Figure~\ref{fig:ckptablate} shows that the BLEU difference between self-training and sentence-level models monotonically increases as the quality of the sentence-level model gets better on the NIST dataset. This implies that we can expect a larger improvement from applying self-training as we improve the sentence-level model on the WMT dataset. Preliminary experiments on training back-translated models didn't improve results on the WMT dataset. We leave further investigation of ways to improve the sentence-level model on the WMT dataset for future work.

The results on OpenSubtitles evaluation sets are in Table~\ref{tab:opusresults}. Our self-training and oracle self-training approaches give the performance improvement of $0.1$ and $0.3$ BLEU respectively. We hypothesize that the small improvement of self-training is due to relatively small number of sentences in the documents in the OpenSubtitles dataset. We validate this claim by varying the number of sentences in the document used for self-training on NIST dataset. Figure~\ref{fig:numsent} shows that the  self-training approach achieves higher BLEU improvement as we increase the number of sentences in documents used for self-training.

\section{Human Evaluation}
\begin{table*}[t!]
\centering
\input{tables/opus-qualitative.tex}
\caption{Four reference documents together with translations generated by the baseline sentence-level model and by our self-training approach. First two documents are taken from the OpenSubtitles English-Russian and second two documents are taken from the NIST Chinese-English dataset.}
\label{tab:opusqualitative}
\end{table*}

We conduct a human evaluation study on the NIST Zh-En and the OpenSubtitles En-Ru datasets. For both datasets we sample 50 documents from the test set where translated documents generated by the self-training approach are not exact copies of the translated documents generated by the sentence-level baseline model. For the NIST Zh-En dataset we present reference documents, translated documents generated by the sentence-level baseline, and translated documents generated by self-training approach to 4 native English speakers. For the OpenSubtitles En-Ru dataset we follow a similar setup, where we present reference documents, translated documents generated by sentence-level baseline, and translated documents generated by self-training approach to 4 native Russian speakers. All translated documents are presented in random order with no indication of which approach was used to generate them.  We highlight the differences between translated documents when presenting them to human evaluators. The human evaluators are asked to pick one of two translations as their preferred option for each document. We ask the human evaluators to consider fluency, idiomaticity and correctness of the translation relative to the reference when entering their preferred choices.

We collect a total of 200 annotations for 50 documents from all 4 human evaluators and show results in Table~\ref{tab:nist_human_eval}. For both datasets, human evaluators prefer translated documents generated by the self-training approach to translated documents generated by the sentence-level model. For NIST Zh-En, 122 out of 200 annotations indicate a preference towards translations generated by self-training approach. For OpenSubtitles En-Ru, 118 out of 200 annotations similarly show a preference towards translations generated by our self-training approach. This is a statistically significant preference $p < 0.05$ according to two-sided Binomial test. When aggregated for each document by majority vote, for NIST Zh-En, translations generated by the self-training approach are considered better in 25 documents, worse in 12 documents, and the same in 13 documents. For OpenSubtitles En-Ru, translations generated by self-training approach are considered better in 23 documents, worse in 15 documents, and the same in 12 documents. The agreement between annotators for NIST Zh-En and OpenSubtitles En-Ru is $\kappa = 0.293$ and $\kappa = 0.320$ according to Fleiss' kappa~\citep{fleiss1971mns}. For both datasets, the inter-annotator agreement rate is considered fair.

\section{Qualitative Results}

In Table~\ref{tab:opusqualitative}, we show four reference document pairs together with translated documents generated by the baseline sentence-level model and by our self-training approach. We emphasize the underlined words in all documents. 

In the first two examples we emphasize the gender of the person marked on verbs and adjectives in translated Russian sentences. In the first example, the baseline sentence-level model inconsistenly produces different gender markings on the underlined verb \foreignlanguage{russian}{сказал} (masculine told) and underlined adjective \foreignlanguage{russian}{сильной} (feminine strong). The self-training approach correctly generates a translation with consistent male gender markings on both the underlined verb \foreignlanguage{russian}{сказал} and the underlined adjective \foreignlanguage{russian}{сильным}. Similarly, in the second example, the baseline model inconsistenly produces different gender markings on the underlined verbs \foreignlanguage{russian}{приглашена} (feminine invited) and \foreignlanguage{russian}{поругался} (masculine fought). Self-training consistently generates female gender markings on both the underlined verbs \foreignlanguage{russian}{приглашена} (feminine invited) and \foreignlanguage{russian}{поссорилась} (feminine fought).

In the third example, we emphasize the underlined named entity in reference and generated translations. The baseline sentence-level model inconsistently generates the names "doyle" and "du" when referring to the same entity across two sentences in the same document. The self-training approach consistently uses the name "doyle" across two sentences when referring to the same entity. In the fourth example, we emphasize the plurality of the underlined words. The baseline model inconsistenly generates both singular and plural forms when referring to same noun in consecutive sentences. Self-training generates the noun "pilots" in correct plural form in both sentences.

\section{Conclusion}
In this paper, we propose a way of incorporating the document context inside a trained sentence-level neural machine translation model using self-training. We process documents from left to right multiple times and self-train the sentence-level NMT model on the pair of source sentence and generated target sentence. This reinforces the choices made by the NMT model thus making it more likely that the choices will be repeated in the rest of the document.

We demonstrate the feasibility of our approach on three machine translation datasets: NIST Zh-En, WMT'19 Zh-En and OpenSubtitles En-Ru. We show that self-training improves sentence-level baselines by up to $0.93$ BLEU. We also conduct a human evaluation study and show a strong preference of the annotators to the translated documents generated by our self-training approach. Our analysis demonstrates that self-training achieves higher improvement on longer documents and using better sentence-level models.

In this work, we only use self-training on source-to-target NMT models in order to capture the target side document context. One extension could investigate the application of self-training on both target-to-source and source-to-target sentence-level models to incorporate both source and target document context into generated translations. Overall, we hope that our work would motivate novel approaches of making trained sentence-level models better suited for document translation at decoding time. 

\section{Acknowledgements}
We would like to thank Phil Blunsom, Kris Cao, Kyunghyun Cho, Chris Dyer, Wojciech Stokowiec and members of the Language team for helpful suggestions.
\bibliography{main}
\bibliographystyle{acl_natbib}

\end{document}

%% file: tables/nist-results.tex
\small
\centering
\begin{tabular}[t]{c|c|c|cccc}
\toprule
Method & Model & MT06 & MT03 & MT04 & MT05 & MT08 \\
\midrule
\citep{wang2017exploitcorss} & RNNSearch & 37.76 & - & - & 36.89 & 27.57 \\
\citep{kuang2017coherencenmt} & RNNSearch & - & - &  38.40 & 32.90 & 31.86 \\
\citep{kuang2017coherencenmt} & Transformer & 48.14 &  48.05 & 47.91 & 48.53 & 38.38 \\
\citep{zhang2018doctransformer} & Doc Transformer &  49.69 & 50.21 & 49.73 & 49.46 & 39.69 \\
\midrule
\multirow{2}{*}{\footnotesize Ours (Non-BT)}
 & Transformer & 48.90 & 48.55
 & 49.55 & 48.04 & 40.89 \\
 & Transformer + self-train & 49.17 & 49.46 & 50.12 & 48.67 & 41.18  \\
\midrule
\multirow{2}{*}{\footnotesize Ours (BT)}
 & Transformer & 51.29 & 51.69
 & 52.48 & 53.00 & 42.72 \\
 & Transformer + self-train & 52.30 & 53.36 & 52.83 & 53.67 & 43.68 \\
\toprule
\end{tabular}
\caption{Results on NIST evaluation sets. The first four rows show the performance of the previous document-level NMT models from ~\citep{wang2017exploitcorss,kuang2017coherencenmt,zhang2018doctransformer}. The last four rows show performance of our baseline sentence-level Transformer models with and without self-training. BT: backtranslation.}

%% file: tables/nist-ablation-results.tex
\small
\centering
\begin{tabular}[t]{lc|c|ccccc}
\toprule
& Model & MT06 & MT03 & MT04 & MT05 & MT08 & $\Delta_{\text{BLEU}}$ \\
\midrule
\multirow{4}{*}{\rotatebox[origin=c]{90}{\footnotesize Non-BT}}
& Baseline & 48.90 & 48.55 & 49.55 & 48.04 & 40.89 & - \\
& ST & 49.26 & 48.91 & 49.81 & 48.27 & 41.10 & +0.28 \\
& Multi-pass ST & 49.17 & 49.46 & 50.12 & 48.67 & 41.18 & +0.53 \\
& Oracle ST & 49.82 & 49.59 & 49.93 & 48.58 & 42.30 & +0.86 \\
\midrule
\multirow{4}{*}{\rotatebox[origin=c]{90}{\footnotesize BT}}
& Baseline & 51.29 & 51.69
 & 52.48 & 53.00 & 42.72 & - \\
& ST & 52.19 & 52.45 & 52.63 & 53.01 & 43.25 & +0.47 \\
& Multi-pass ST & 52.30 & 53.36 & 52.83 & 53.67 & 43.68 & +0.93 \\
& Oracle ST & 53.11 & 53.90 & 53.05 & 53.81 & 45.45 & +1.63 \\
\toprule
\end{tabular}
\caption{Ablation study on NIST evaluation sets measuring the effect on multiple passes of decoding and the oracle on self-training procedure. BT: backtranslation. ST: self-training.}

%% file: tables/wmt-results.tex
\small
\centering
\begin{tabular}[t]{c|c|c|c|cccc}
\toprule
Method & Architecture & Model & Valid17 & Test17 & Test18 & Test19 & $\Delta_{\text{BLEU}}$ \\
\midrule
\citep{xia2019wmtmicrosoft} & Transformer Big & - & - & 24.2 & 24.5 & - & - \\
\midrule
\multirow{4}{*}{Ours} & \multirow{4}{*}{Transformer Big} & Baseline & 22.5 & 23.4 & 24.1 & 25.4 & - \\
& & ST & 22.6 & 23.4 & 24.1 & 25.5 & +0.0 \\
& & Multi-pass ST & 22.6 & 23.5 & 24.2 & 25.6 & +0.1 \\
& & Oracle ST & 24.3 & 25.7 & 25.6 & 29.8 & +2.5 \\
\bottomrule
\end{tabular}
\caption{Results on WMT'19 Chinese-English evaluation sets. The first row shows the performance of the Transformer Big model by ~\cite{xia2019wmtmicrosoft}. All models were trained without additional monolingual data and without pretraining. ST: self-training. }

%% file: tables/opus-results.tex
\small
\centering
\begin{tabular}[t]{lc|ccc}
\toprule
& Model & Valid & Test & $\Delta_{\text{BLEU}}$ \\
\midrule
\multirow{4}{*}{\rotatebox[origin=c]{90}{\footnotesize Non-BT}}
& Baseline & 31.38 & 31.38 & - \\ 
& ST & 31.41 & 31.39 & +0.02  \\
& Multi-pass ST & 31.44 & 31.43 & +0.06 \\
& Oracle ST & 31.65 & 31.66 & +0.28 \\
\midrule
\multirow{4}{*}{\rotatebox[origin=c]{90}{\footnotesize BT}}
& Baseline & 32.22 & 32.21 & - \\ 
& ST & 32.24 & 32.23 & +0.02 \\
& Multi-pass ST & 32.25 & 32.26 & +0.04 \\
& Oracle ST & 32.56 & 32.54 & +0.34 \\
\bottomrule
\end{tabular}
\caption{Results on OpenSubtitles English-Russian evaluation sets. ST: self-training.}

%% file: tables/nist-human-eval.tex
\small
\centering
\begin{tabular}[t]{c|c|cc}
\toprule
Dataset & Total & Self-train & Baseline \\
\midrule
NIST Zh-En & 200 & 122 & 78 \\
\midrule
OpenSubtitles En-Ru & 200 & 118 & 82 \\
\end{tabular}

%% file: tables/opus-qualitative.tex
\resizebox{\textwidth}{!}{
\begin{tabular}{ll}
    \midrule
    {\bf Ref}        &  \foreignlanguage{russian}{мы с эйприл развелись . как я и \underline{сказал} ... игра в ожидание . будь \underline{сильным} . и всё получится .}\\
    {\bf Baseline} & \foreignlanguage{russian}{мы с эйприл развелись . ну , как я уже \underline{сказал} ... игра ожидания . будь \underline{сильной} . ты справишься .} \\
    {\bf Ours} & \foreignlanguage{russian}{мы с эйприл развелись . ну , как я уже \underline{сказал} ... игра ожидания . будь \underline{сильным} . ты справишься .} \\
    \bottomrule
    \end{tabular}
}
\resizebox{\textwidth}{!}{
\begin{tabular}{ll}
    \midrule
    {\bf Ref}        &  \foreignlanguage{russian}{сёрен устраивает вечеринку по поводу своего дня рождения в субботу , а я не знаю , \underline{пойду} ли я .}\\
    & \foreignlanguage{russian}{почему бы тебе не пойти ? просто всё пошло не так . - и я \underline{поссорился} с кнудом .} \\
    {\bf Baseline} & \foreignlanguage{russian}{в субботу день рождения сёрена и я не знаю , \underline{приглашена} ли я . } \\
    & \foreignlanguage{russian}{почему тебя не пригласили ? все просто пошло не так . - и я \underline{поругался} с кнудом .} \\
    {\bf Ours} & \foreignlanguage{russian}{в субботу день рождения сёрена и я не знаю , \underline{приглашена} ли я . } \\
    & \foreignlanguage{russian}{почему тебя не пригласили ? все просто пошло не так . - и я \underline{поссорилась} с кнудом .} \\
    \bottomrule
    \end{tabular}
}
\resizebox{\textwidth}{!}{
\begin{tabular}{ll}
    \midrule
    {\bf Ref}        &  we are actively seeking a local partner to set up a joint fund company , " \underline{duchateau} said . \\
    &  \underline{duchateau} said that the chinese market still has ample potentials . \\
    {\bf Baseline}        & we are actively looking for a local partner to establish a joint venture fund company , " \underline{doyle} said .\\
    & \underline{du} said that there is still a lot of room for the chinese market . \\
    {\bf Ours}        &  we are actively looking for a local partner to establish a joint venture fund company , " \underline{doyle} said .\\
    & \underline{doyle} said that there is still great room for the chinese market . \\
    \bottomrule
    \end{tabular}
}
\resizebox{\textwidth}{!}{
\begin{tabular}{ll}
    \midrule
    {\bf Ref}        &  in may this year , \underline{13 pilots} with china eastern airlines wuhan company \\
    & in succession handed in their resignations , which were rejected by the company . \\
    & soon afterwards , \underline{the pilots} applied one after another at the beginning of june \\
    & to the labor dispute arbitration commission of hubei province for labor arbitration , \\
    & requesting for a ruling that their labor relationship with china eastern airlines wuhan company be terminated . \\
    {\bf Baseline} & in may this year , \underline{13 pilots} of china eastern 's wuhan company \\
    & submitted their resignations one after another , but the company refused .	\\
    & \underline{the pilot} then applied for labor arbitration with the hubei province \\
    & labor dispute arbitration committee in early june , requesting the ruling \\
    & to terminate the labor relationship with the wuhan company of china eastern airlines . \\
    {\bf Ours} & in may this year , \underline{13 pilots} of china eastern 's wuhan company \\
    & submitted their resignations one after another , but the company refused . \\
    & subsequently , in early june , \underline{the pilots} successively applied for labor arbitration \\
    & with the hubei province labor dispute arbitration committee , \\
    & requesting that the labor relationship with china eastern airlines be terminated .\\
    \bottomrule
    \end{tabular}
}